\documentclass[10pt,twocolumn,letterpaper]{article}

\usepackage[pagenumbers]{cvpr} 

\definecolor{cvprblue}{rgb}{0.21,0.49,0.74}
\usepackage[pagebackref,breaklinks,colorlinks,allcolors=cvprblue]{hyperref}

\usepackage{pifont}
\usepackage{multirow}
\usepackage{wrapfig}
\usepackage{verbatim}

\def\paperID{14639} 
\def\confName{CVPR}
\def\confYear{2025}

\title{LoRA of Change: Learning to Generate LoRA for the Editing Instruction from A Single Before-After Image Pair}

\author{
Xue Song\textsuperscript{1}, Jiequan Cui\textsuperscript{2}, Hanwang Zhang\textsuperscript{2}, Jiaxin Shi\textsuperscript{3}, Jingjing Chen\textsuperscript{1},
Chi Zhang\textsuperscript{4}\thanks{This work was advised by Chi
Zhang and supported by Westlake University.}, Yu-Gang Jiang\textsuperscript{1}\\
{\small \textsuperscript{1}Shanghai Key Lab of Intell. Info. Processing, School of CS, Fudan University} \\
{\small \textsuperscript{2}Nanyang Technological University \qquad \textsuperscript{3}Xmax.AI \qquad \textsuperscript{4}Westlake University} \\
{\tt\footnotesize \{xsong18, chenjingjing, ygj\}@fudan.edu.cn, jiequancui@gmail.com, hanwangzhang@ntu.edu.sg,} \\
{\tt\footnotesize shijx12@gmail.com, chizhang@westlake.edu.cn}}

\begin{document}
\maketitle

\begin{abstract}
In this paper, we propose the LoRA of Change (LoC) framework for image editing with visual instructions, \ie, before-after image pairs. Compared to the ambiguities, insufficient specificity, and diverse interpretations of natural language, visual instructions can accurately reflect users' intent. 
Building on the success of LoRA in text-based image editing and generation, we dynamically learn an instruction-specific LoRA to encode the “change” in a before-after image pair, enhancing the interpretability and reusability of our model.
Furthermore, generalizable models for image editing with visual instructions typically require quad data, \ie, a before-after image pair, along with query and target images. Due to the scarcity of such quad data, existing models are limited to a narrow range of visual instructions. To overcome this limitation, we introduce the LoRA Reverse optimization technique, enabling large-scale training with paired data alone. Extensive qualitative and quantitative experiments demonstrate that our model produces high-quality images that align with user intent and support a broad spectrum of real-world visual instructions.

\end{abstract}    
\section{Introduction}
\label{sec:intro}
\begin{figure}
    \centering
    \includegraphics[width=0.95\linewidth]{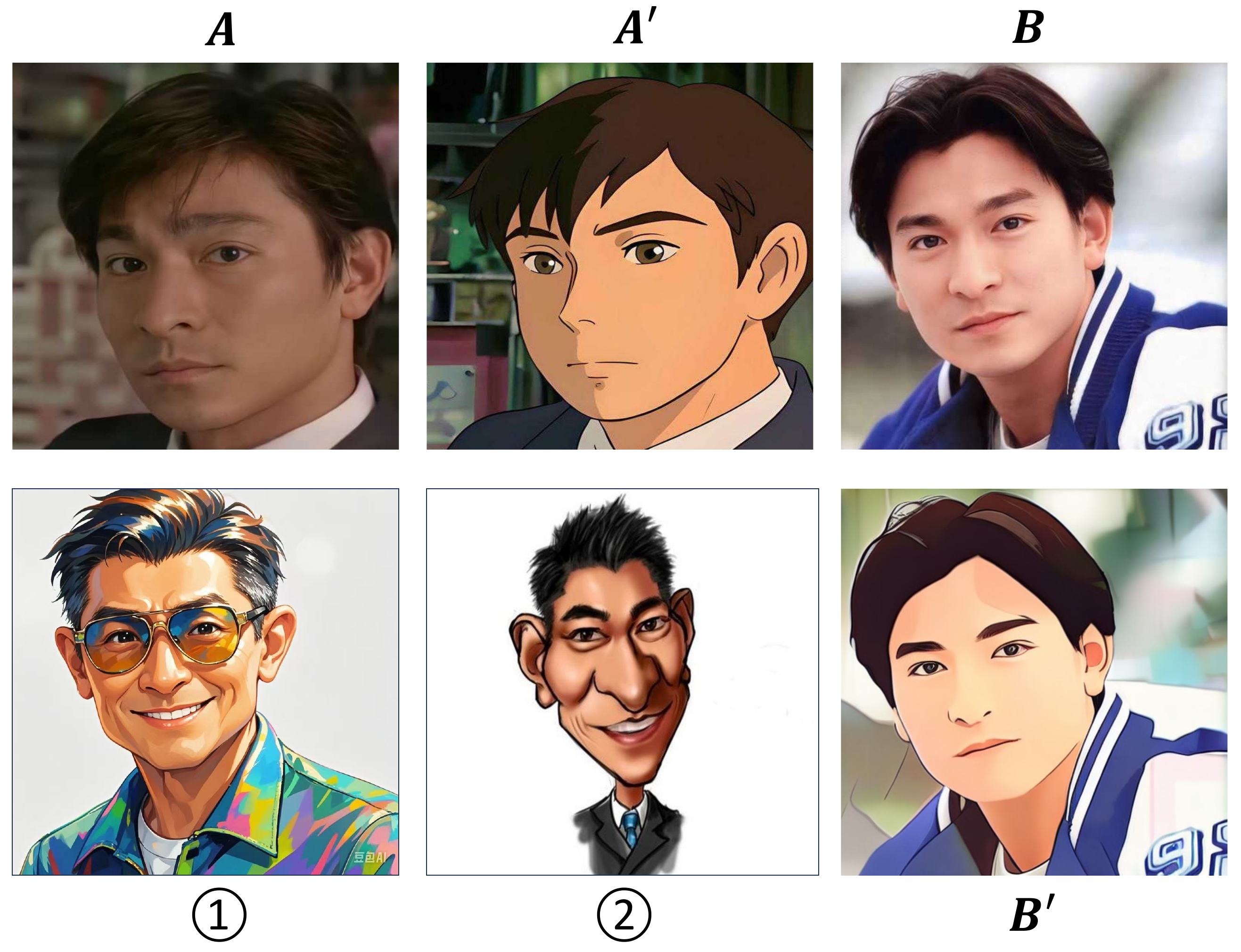}
    \caption{Image editing with before-after image pair instructions.}
    \label{fig:before_after_pair}
\end{figure}

Text-based image editing modifies real images based on language instructions. However, these instructions can sometimes fail to accurately convey users' intentions due to ambiguities, insufficient specificity, and diverse interpretations of natural language. To address these limitations, recent works~\cite{gu2024analogist,yang2024imagebrush,bar2022visual} explore the use of visual instructions, such as before-after image pairs, to better capture the user's intent. As shown in Figure~\ref{fig:before_after_pair}, given a before-after image pair $<A, A^{'}>$ as an editing instruction, the user expects to derive the edited image $B^{'}$ with a query image $B$. Obviously, it is hard to describe the desired editing as a sentence, like ``A cartoon picture of Andy Lau", because images \ding{192} and \ding{193} are also aligned with the text prompt.

Existing image editing methods that leverage visual instructions can be categorized into three primary approaches. One approach~\cite{bar2022visual, gu2024analogist,yang2024imagebrush} reformulates image editing as an inpainting task. Given a set of training examples $<A, A^{'}, B, B^{'}>$, where $<A, A^{'}>$ represents the before-after image pair, $B$ and $B^{'}$ are the query and target images respectively, the images are arranged into a grid-like composite. Then, models are trained with inpainting-related techniques, like masked image modeling. In contrast, other methods~\cite{wang2023context, meng2024instructgie} employ ControlNet \cite{zhang2023adding}, taking $A, A^{'}$ and $B$ as spatial conditions. 
Both approaches suffer from the scarcity of \textit{quad data} and fail to explicitly extract editing instructions from before-after image pairs, thus leading to limited interpretability and reusability. A third approach encodes visual instructions using the textual inversion technique~\cite{nguyen2023visual}. However, due to the limited representational capacity of text, this method struggles to capture fine-grained visual instructions. Additionally, the tuning process for instruction inversion is often time-consuming.

\noindent{\bf Our Algorithm --- LoRA of Change.} To tackle the above-mentioned issues, we propose the \textit{LoRA of Change (LoC)} framework as shown in Figure~\ref{fig:framework}. Considering the success of LoRA in text-based image editing~\cite{Song_2024_CVPR,gandikota2023concept,chen2024fastedit} and customized image generation~\cite{gu2024mix,shah2025ziplora,wu2024mixture}, we learn to dynamically generate an \textit{instruction-specific} LoRA for each before-after image pair, which means the desired editing instruction is explicitly extracted into LoRA weights. 
For the LoRA generation, we design a hypernetwork $\mathcal{H}$ which takes $<A, A^{'}>$ as inputs and outputs a LoRA. 
Applying the LoRA to the frozen InstructPix2Pix~\cite{brooks2023instructpix2pix} model $\mathcal{G}$ with the query image $B$ as the spatial condition, our model is optimized to reconstruct the target image $B^{'}$:
\begin{equation}
    \mathcal{H}=\arg \min_{\mathcal{H}} ||\mathcal{G}(\mathcal{H}(A, A^{'}), B) - B^{'}||.
    \label{eq:lora_change_1}
\end{equation}
At inference, with a before-after image pair $<A, A^{'}>$ and a query image $B$, we follow the generation process of DDIM~\cite{song2020denoising} to derive the edited image $B^{'}=\mathcal{G}(\mathcal{H}(A, A^{'}), B)$. 

\begin{figure}
    \centering
    \includegraphics[width=1.0\linewidth]{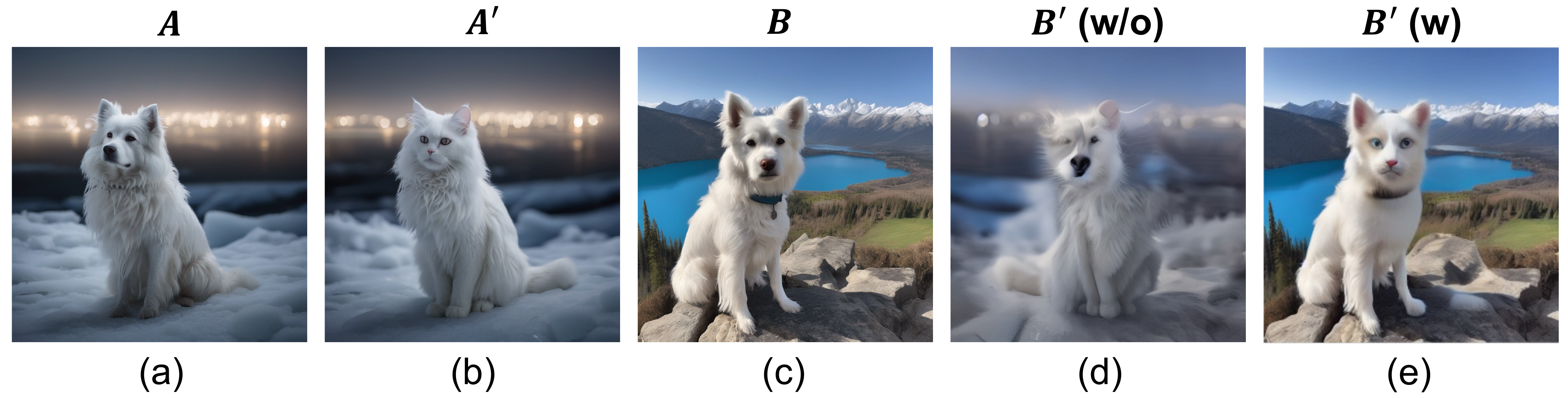}
    \caption{Information leakage without \textit{LoRA Reverse} training.}
    \label{fig:leakage}
\end{figure}

\noindent{\bf LoRA Reverse Training.}
Due to the scarcity of \textit{quad data}, prior work~\cite{wang2023images,bar2022visual,wang2023context} has focused mainly on specific visual instructions, such as transformations between natural images and segmentation maps, depth maps, or edge maps. Consequently, these trained models are unable to handle diverse visual instructions in real-world applications.
To enable the large-scale training with various visual instructions, we only require \textit{paired data}, \ie, \textit{we set $B$ and $B^{'}$ as the randomly horizontal flip of $A$ and $A^{'}$ respectively}. 
Under this setting, an information leakage problem arises: 
\textit{The hypernetwork $\mathcal{H}$ is designed to generate instruction-specific LoRA. However, the generation model $\mathcal{G}$ can successfully reconstruct $B^{'}$ even if the LoRA only encodes the $A^{'}$ without focusing on the ``change'' between $A$ and $A^{'}$.} As illustrated in Figure~\ref{fig:leakage}(d), we observe that visual appearance information can inadvertently leak into the LoRA generation process, resulting in the failure of image editing.

To address the issue, we develop a \textit{LoRA Reverse} technique to regularize the optimization process. Our key insight is that the extracted \textit{instruction-specific} LoRA encodes the visual instruction, and reversing the LoRA should enable editing $B^{'}$ back to $B$. Thus, besides Eq.~\eqref{eq:lora_change_1}, we add the following training objective,
\begin{equation}
    \mathcal{H}=\arg \min_{\mathcal{H}} ||\mathcal{G}(-\mathcal{H}(A, A^{'}), B^{'}) - B||.
    \label{eq:lora_change_2}
\end{equation}
In the course of training, we also randomly exchange from $<A, A^{'}>$ to $<A^{'}, A>$, implicitly to encourage the consistency of $\mathcal{H}(A, A^{'}) = -\mathcal{H}(A^{'}, A)$. As shown in Figure~\ref{fig:leakage}(e), the appearance leakage issue is greatly alleviated with the \textit{LoRA Reverse} training.  

\begin{figure*}[!htp]
    \centering
    \includegraphics[height=1.0\textheight]{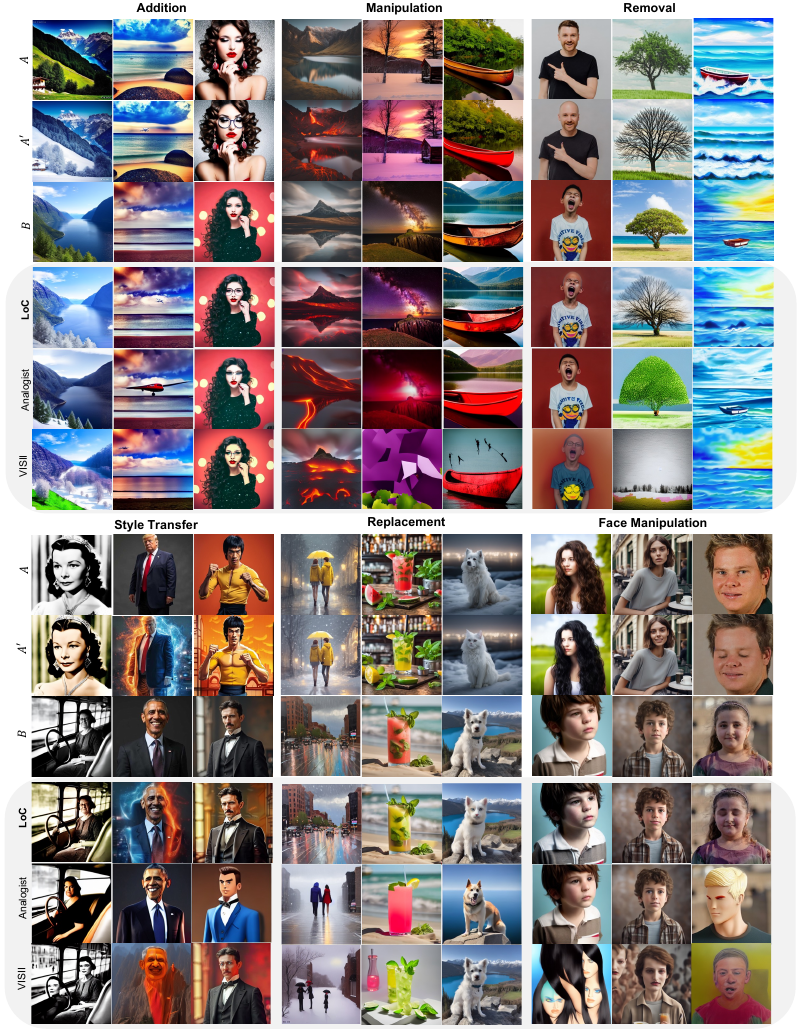}     
    \caption{Comparison of qualitative examples across the 6 editing types between our LoC and two SOTAs.}
\label{fig:visual_comparisions}
\end{figure*}

Finally, we perform experiments on SEED-Data-Edit~\cite{ge2024seed} and MagicBrush~\cite{zhang2024magicbrush} datasets. As illustrated in Figure~\ref{fig:visual_comparisions}, our approach achieves a significantly improved quality of edited images,
keeping better fidelity and being more aligned with the before-after image pairs when compared to prior methods. 
Leveraging large-scale \textit{paired} training data, our model support a wide range of visual instructions, including \textit{addition, manipulation, removal, style transfer, replacement, and face manipulation}. Notably, our method enables real-time image editing without the need for test-time finetuning.

\noindent{\bf Summary.}
Overall, our primary contribution is the proposed \textit{LoRA of Change (LoC)} framework, which encodes the ``change'' in before-after image pairs into dynamic LoRAs. Additionally, the designed \textit{LoRA Reverse} training effectively mitigates the problem of appearance leakage from before-after image pairs, enabling large-scale training on \textit{paired data} and thus supporting various visual instructions in the real world.

\section{Related Work}
\label{sec:relatedwork}

\begin{table*}[t]
    \centering
    \caption{Comparison with existing methods for image editing with visual instructions. Analogist and VISII do not require large-scale training. Our method is only trained on \textit{paired data}.}
    \resizebox{1.0\linewidth}{!}
    {
    \begin{tabular}{ccccc}
         \toprule
         Methods  &Training Data  &Interpretability & Extra assistant &Support editing \\
         \midrule
         Visual Prompting \cite{bar2022visual} &Quad data &\ding{55} &None &\multirow{3}{*}{\shortstack{\ding{192} Tasks that ground truth is easy to construct, like segmentation, colorization, edge detection, \\ style transfer, depth estimation, or low-level vision tasks including image denoising.}} \\
         Painter \cite{wang2023images}        &Quad data &\ding{55} &None & \\
         PromptGIP \cite{liu2023unifying}       &Quad data &\ding{55} &None & \\
         \cmidrule{4-5}
         Imagebrush \cite{yang2024imagebrush}      &Quad data &\ding{55} &Bounding Box &\multirow{2}{*}{Some tasks of \ding{192}; \ding{193} image editing.}\\
         Analogist \cite{gu2024analogist}       &N/A &\ding{55} &GPT-4V & \\
         \midrule
         InstructGIE \cite{meng2024instructgie}     &Quad data &\ding{55} &Text prompt &Some tasks of \ding{192}; \ding{193} image editing.\\
         Prompt Diffusion \cite{wang2023context} &Quad data &\ding{55} &Text prompt &Some tasks of \ding{192}.\\
         \midrule
         VISII \cite{nguyen2023visual}           &N/A &\ding{51} &Test time fine-tuning &Some tasks of \ding{192}; \ding{193} image editing. \\
         \midrule
         LoC (Ours)       &Paired data &\ding{51} &None &Some tasks of \ding{192}; \ding{193} image editing. \\
         \bottomrule
    \end{tabular}
    }
    \label{tab:related_work}
\end{table*}

\subsection{Text-based Image Editing}

Text-based image editing requires keeping the fidelity of the source image, as well as realizing editability aligned with text prompts. Previous works could be divided into two groups: \textit{1) methods trained on large-scale paired data with language editing instructions and 2) zero-shot ones utilizing prior knowledge in foundation models.} 

The former type \cite{brooks2023instructpix2pix,huang2024smartedit, Sheynin2023EmuEP} trains a task-specific model with large-scale paired data and language editing instructions, achieving image editing by the generalization ability of the trained models on unseen data. 
The zero-shot approaches edit images by utilizing the prior knowledge of generative models (e.g., Stable Diffusion~\cite{rombach2022high}). 
This kind of method often adheres to the paradigm of reconstruction then followed by editing. For the reconstruction stage, tuning-based methods fine-tune the generative models, e.g., UNet~\cite{kawar2023imagic,zhang2023sine}, LoRA~\cite{Song_2024_CVPR}, while inversion-based ones adopt DDIM inversion \cite{song2020denoising} or its advanced versions~\cite{pan2023effective,wallace2023edict,garibi2024renoise}. As for the editing stage, there are various strategies to boost editing performance, e.g., reforming images as editable elements~\cite{mu2024editable}, explicitly extracting semantic changes~\cite{Song_2024_CVPR}, and adapting self-attention and cross-attention maps~\cite{hertz2022prompt, Tumanyan_2023_CVPR}.

\subsection{Vision-based Image Editing}
Vision-based image editing refers to performing edits using visual instructions, where each instruction consists of a before-and-after image pair that illustrates the changes induced by the instruction. Previous methods can be categorized into the following three kinds.

\textit{1) Inpainting-based methods} reframe this task as image inpainting. The approaches \cite{gu2024analogist,yang2024imagebrush,bar2022visual} often construct a $2 \times 2$ grid image using $A$, $A'$, $B$, and a masked image, and then models are optimized to reconstruct the masked image with $B'$ as the ground truth. Particularly, to fully exploit the visual contextual information, Analogist \cite{gu2024analogist} utilizes the self-attention map between $A$ and $A'$ to guide the change from $B$ to $B'$ while ImageBrush \cite{yang2024imagebrush} carefully designs an encoder regarding $A$, $A'$, and $B$ as prompts. 
Additionally, works~\cite{wang2023images,liu2023unifying} train models with masked image modeling (MIM), which randomly masks patches in the grid image during training and masks the patches in the fourth sub-grid during inference. 
\textit{2) ControlNet-based methods.} Derived from the success of ControlNet~\cite{zhang2023adding} in the conditional image generation, works~\cite{wang2023context, meng2024instructgie} utilize this powerful model by taking $A$, $A'$, and $B$ as spatial conditions. 
\textit{3) Methods based on textual inversion.} VISII~\cite{nguyen2023visual} fulfills visual instruction inversion for a single example via optimizing a part of text embedding. However, it suffers from the limited expressiveness of such text-related spaces and the time-consuming nature of single-example inversion tuning.
Systematic comparisons between our \textit{LoC} method and previous ones are shown in Table~\ref{tab:related_work}.

\section{Our Method --- LoRA of Change}

\begin{figure*}[t]
    \centering
    \includegraphics[width=1.0\linewidth]{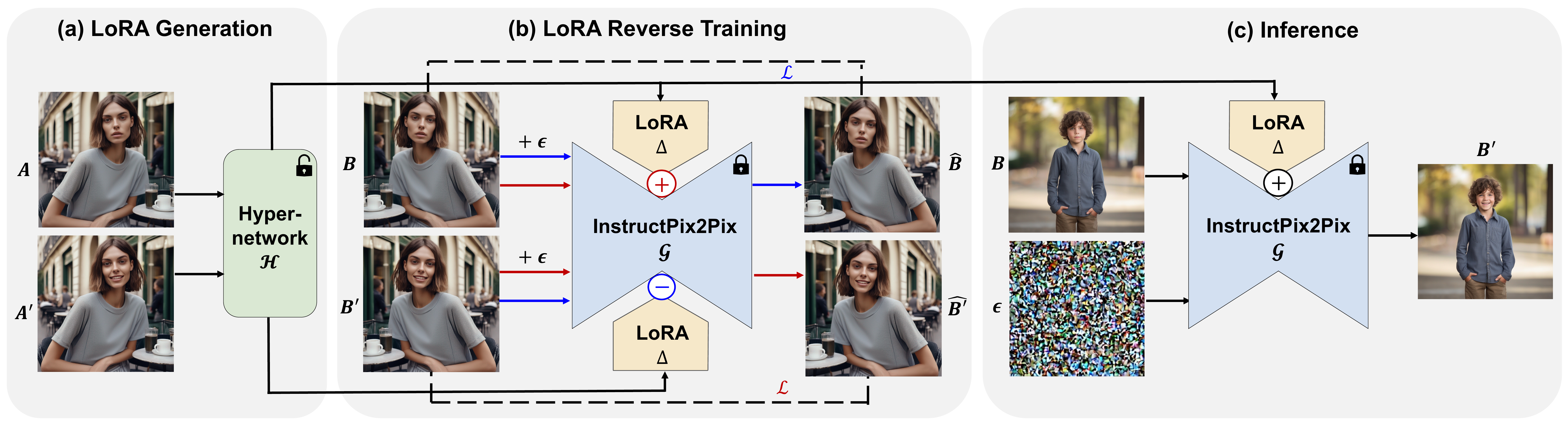}
    \caption{Overview of \textit{LoRA of Change (LoC)} framework. (a) shows that the hypernetwork $\mathcal{H}$ generates the \textit{instruction-specific} LoRA with the before-after image pair $<A, A^{'}>$ as inputs. (b) presents the LoRA reverse training. 
    With the generated LoRA $\Delta$, 
    the \textcolor{red}{red arrows $\to$} indicate that the model is trained to reconstruct $B^{'}$ taking $B$ as spatial condition while the \textcolor{blue}{blue arrows $\to$} indicate that the model is trained to reconstruct $B$ taking $B^{'}$ as spatial condition. (c) is the inference for image editing. $\mathcal{L}$ is the image reconstruction loss.}
    \label{fig:framework}
\end{figure*}

\begin{figure}[t]
    \centering
    \includegraphics[width=1.0\linewidth]{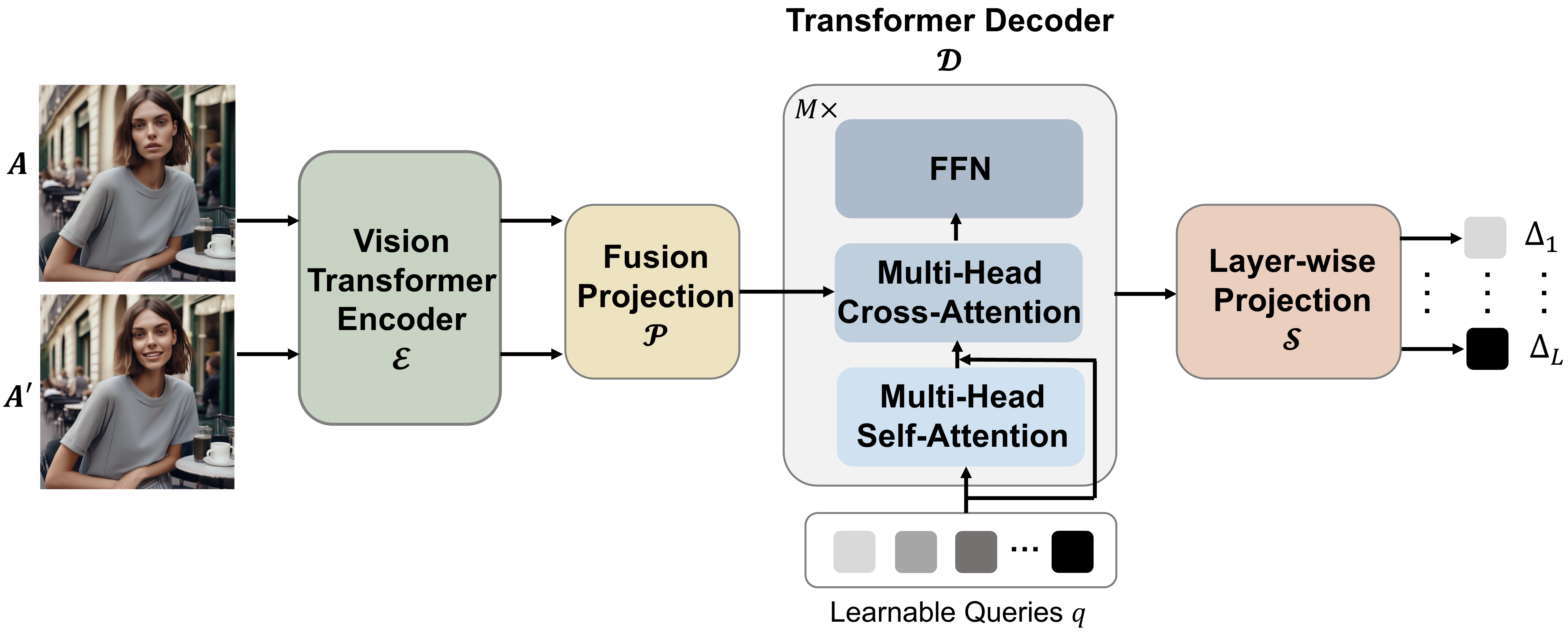}
    \caption{Hypernetwork $\mathcal{H}$ for LoRA generation. The transformer decoder $\mathcal{D}$ consists of $M=6$ blocks.}
    \label{fig:hypernetwork}
\end{figure}

\label{sec:method}
In this section, we detail the implementation of our \textit{LoRA of Change (LoC)} framework. As shown in Figure~\ref{fig:framework}, \textit{LoC} consists of a hypernetwork $\mathcal{H}$ to generate \textit{instruction-specific} LoRAs, and an InstructPix2Pix generation model $\mathcal{G}$ for image generation.   
Regarding LoRA generation, the hypernetwork design and weight initialization are introduced in Section~\ref{sec:lora_gen}. Then, we discuss the optimization of \textit{LoRA Reverse} in Section~\ref{sec:lora_reverse}.

\subsection{LoRA Generation}
\label{sec:lora_gen}
LoRA~\cite{hu2021lora} has been a critical component for finetuning foundation models in downstream tasks. Particularly, it finds various applications in text-based image editing and customized image generation. In this case, LoRA is optimized to encode a scene~\cite{shah2025ziplora,gu2024mix} or an object~\cite{gu2024mix,shah2025ziplora,wu2024mixture}. In this paper, we explore the use of dynamic LoRAs to encode visual instructions.

\noindent{\bf Hypernetwork Design.}
Our hypernetwork $\mathcal{H}$ consists of a ViT~\cite{alexey2020image} encoder $\mathcal{E}$ that takes the before-after image pair $<A, A^{'}>$ as inputs, a linear layer $\mathcal{P}$ to transform image-level features to the visual instruction feature, a transformer decoder $\mathcal{D}$ that outputs primitive features for LoRA, and then a set of linear layers $\mathcal{S}$ to project primitive features into LoRA weights for the visual instruction $<A, A^{'}>$.

Specifically, with the ViT encoder $\mathcal{E}$, we derive the image-level features of $A$ and $A^{'}$, and concat them together, then followed by a linear fusion layer $\mathcal{P}$ to output the visual instruction feature $f_{vis\_ins}$,
\begin{eqnarray}
    f_{A}, f_{A^{'}} &=& \mathcal{E}(A), \mathcal{E}(A^{'}), \\
    f_{vis\_ins} &=& \mathcal{P}(concat(f_{A}, f_{A^{'}})),
\end{eqnarray}
where $f_{A}, f_{A^{'}} \in \mathbb{R}^{196 \times 768}$, $f_{vis\_ins} \in \mathbb{R}^{196 \times 512}$, $\mathcal{P}$ is a linear layer. 

The visual instruction feature $f_{vis\_ins}$ is then fed into the decoder network $\mathcal{D}$. As illustrated in Figure~\ref{fig:hypernetwork}, the decoder $\mathcal{D}$ is composed of learnable queries $q \in \mathbb{R}^{32 \times 512}$, self-attention layers, cross-attention layers, and feed-forward (FFN) layers. Taking learnable queries $q$ and visual instruction feature $f_{vis\_ins}$ as inputs, the decoder $\mathcal{D}$ output primitive features for LoRA. There are a total of $L=32$ self-attention and cross-attention layers in the UNet of InstructPix2Pix model $\mathcal{G}$. For each of these layers, we use a linear layer to transform the primitive feature into LoRA weights with proper dimensions:
\begin{eqnarray}
    \Delta_{i} = \mathcal{S}[i](\mathcal{D}(q, f_{vis\_ins})[i]),
    \label{eq:i_lora}
\end{eqnarray}
where $i$ is the layer index.

To this end, we derive the \textit{instruction-specific} LoRA $\Delta$ for the before-after image pair $<A, A^{'}>$.

\noindent{\bf Weight Initialization.}
For the ViT encoder, we load the pre-trained weights on ImageNet. The linear fusion layer $\mathcal{P}$ and the decoder $\mathcal{D}$ are randomly initialized. For the initialization of linear layers $S$, we guarantee that the generated LoRA should be zero at the beginning for fast training convergence, which is aligned with the initialization of LoRA in the original paper~\cite{hu2021lora}.


\subsection{LoRA Reverse}
\label{sec:lora_reverse}
Considering the scarcity of \textit{quad data}, we enable the large-scale training with \textit{only paired data} via 
the \textit{LoRA Reverse} optimization shown in Figure~\ref{fig:framework} (b), realizing image editing with a wide spectrum of visual instructions in the real world. The detailed implementations of the \textit{LoRA Reverse} training and editing inference procedure are presented as follows.

\noindent \textbf{LoRA Reverse Training.}
The InstructPix2Pix model~\cite{brooks2023instructpix2pix} supports the spatial condition with an extra image $B$, which is well-aligned with our formulations in Eqs.~\eqref{eq:lora_change_1} and ~\eqref{eq:lora_change_2}. Thus, we use it as our generation model $\mathcal{G}$. Given the training sample set $<A, A^{'}, B, B^{'}>$, especially $B$ and $B^{'}$ are the randomly horizontal flip of $A$ and $A^{'}$,
to optimize our reconstruction objectives of Eqs.~\eqref{eq:lora_change_1} and ~\eqref{eq:lora_change_2}, we solve the following diffusion loss,
\begin{eqnarray}
    \arg \min_{\mathcal{H}} &\mathbb{E}_{(t, \epsilon)}||B^{'} - \Theta(x_{t}, t, B, \mathcal{H}(A, A^{'}))||_{2}^{2} + \label{eq:lora_reverse_1} \\ 
    &\mathbb{E}_{(t, \epsilon)}||B - \Theta(x_{t}, t, B^{'}, -\mathcal{H}(A, A^{'}))||_{2}^{2} \label{eq:lora_reverse_2},
\end{eqnarray}
where $\epsilon \in \mathcal{N}(0, \mathbf{I})$, $t \in [0, T]$ is a sampled time step (T is the maximum), $\Theta(\cdot)$ is the pre-trained $x_{0}$ prediction UNet that is frozen, $x_{t}=\sqrt{\alpha_{t}}x_{0} + \sqrt{1-\alpha_{t}}\epsilon$ is the noisy input at $t$, in particular $x_{0}=B^{'}$ for Eq.~\eqref{eq:lora_reverse_1} while $x_{0}=B$ for Eq.~\eqref{eq:lora_reverse_2}, $B$ in Eq.~\eqref{eq:lora_reverse_1} and $B^{'}$ in Eq.~\eqref{eq:lora_reverse_2} are spatial conditions, and $\alpha_{t}$ is related to a fixed noisy schedule~\cite{song2020denoising, ho2020denoising}.

With the \textit{LoRA Reverse} training objectives in Eqs.~\eqref{eq:lora_reverse_1} and~\eqref{eq:lora_reverse_2}, the model is forced to learn the ``change'' in the before-after image pair $<A, A^{'}>$ rather than just memorizing $A^{'}$, making it feasible to train models with only \textit{paired data}. Compared to \textit{quad data}, \textit{paired data} is easier to collect. Thus, we can achieve large-scale training and then realize a wide spectrum of visual instructions. 

Additionally, we adopt a two-stage training pipeline. In stage-1, we pre-train the model on a large hybrid dataset SEED-Data-Edit~\cite{ge2024seed} consisting of both synthesized data and manually annotated data. In stage-2, we fine-tune the model on a smaller, more curated dataset MagicBrush~\cite{zhang2024magicbrush} and the human-annotated part of SEED-Data-Edit for refinement.

\noindent \textbf{Editing at Inference.}
After training convergence, we follow the generation process of DDIM~\cite{song2020denoising} to edit images.
Given a single before-after image pair $<A,A^{'}>$ and a query image $B$, with a sampled $x_{T} \in \mathcal{N}(0, \mathbf{I})$, we generate the edited image $B^{'}$ with the following iterative update from $t=T$ to $t=0$:
\begin{eqnarray}
    x_{t-1} =& \sqrt{\alpha_{t-1}} \Theta(x_{t},t,B, \mathcal{H}(A,A^{'})) \nonumber \\ 
    &+ \sqrt{1-\alpha_{t-1}} (\frac{x_{t}-\sqrt{\alpha_{t}}\Theta(x_{t},t,B,\mathcal{H}(A,A^{'}))}{\sqrt{1-\alpha_{t}}}),
\end{eqnarray}
where we obtain the edited image $B^{'}=x_{0}$.

\section{Experiment}
\label{sec:exp}

\subsection{Experimental Settings}
\noindent{\bf Datasets.}
Our work leverages the dataset of SEED-Data-Edit~\cite{ge2024seed} and MagicBrush~\cite{zhang2024magicbrush}.
SEED-Data-Edit is developed for text-based image editing. It is a hybrid dataset made up of 3 parts:
1) Sythesized data with diverse image editing pairs using an automatic pipeline; 2) Real-world scenario data scraped from the internet; 3) High precision multi-turn editing data annotated by humans. There are a total of 3.7M image pairs. Compared with the data used for InstructPix2Pix, SEED-Data-Edit is of much higher quality. MagicBrush is a pure manually annotated dataset for real image editing. It consists of 10K image pairs. 
We first use the SEED-Data-Edit data to pre-train our model for 80 epochs. Then, we finetune it with the MagicBrush dataset and manually annotated samples in SEED-Data-Edit for 80 epochs in stage-2. 

The test images of our work are partly from existing datasets, \ie, InstructPix2Pix \cite{brooks2023instructpix2pix}, InstructBrush \cite{zhao2024instructbrush}, InstructGIE \cite{meng2024instructgie}, and DAC \cite{Song_2024_CVPR}, and partly from Unsplash (https://unsplash.com/).

\noindent{\bf Implementation Details.}
For hypernetwork $\mathcal{H}$, we set the number of blocks as $M=6$ and the number of heads in multi-head self/cross-attention as $H=8$ in the decoder $\mathcal{D}$.
The InstructPix2Pix pre-trained model is used as our generation model $\mathcal{G}$.
All images are with $512 \times 512$.
Our models are trained on 8 Tesla H800 GPUs for 80 epochs using the AdamW optimizer. The learning rate is set to 1e-5, and the batch size is set to 96.

\noindent{\bf Comparison Methods.}
To thoroughly evaluate the effectiveness of our method, we compare it with other open-sourced state-of-the-art models tailored for image editing with visual instructions: 1) Analogist, a training-free algorithm that utilizes in-context learning ability of Stable Diffusion models; 2) VISII, a test time fine-tuning method leveraging text inversion techniques; 
3) PromptGIP, an inpainting-based method trained on \textit{quad} data with masked image modeling pipeline. 

It is worth noting that PromptGIP requires \textit{quad data} for training. Thus, the model mainly concentrates on tasks, like image segmentation, colorization, edge detection, depth estimation, and low-level vision tasks including image denoising. 
The ground truths for these tasks can be obtained by task-specific expert models. \textit{However, for image editing including object addition/removal/replacement, and manipulation, there is still no such a strong expert model to generate a large amount of high-quality \textit{quad data}.}

\subsection{Qualitative Evaluation}
We demonstrate the advantages of the proposed \textit{LoC} method with qualitative evaluations.
Specifically, we test our model on 6 types of visual instructions.

\noindent{\bf Wide Spectrum of Visual Instructions.}
We show that our \textit{LoC} model supports a wide spectrum of visual instructions including 1) addition, 2) manipulation, 3) removal, 4) style transfer, 5) replacement, and 6) face manipulation. The experimental results are listed in Figure~\ref{fig:visual_comparisions} and more results are included in the Appendix. For each kind of visual instruction, we provide three image examples. Take an example of face manipulation, given a before-after image pair that ``a smiling man with opened eyes $\to$ the smiling man with closed eyes'' and a query image of a little girl with opened eyes, our model can accurately capture the visual instruction of ``closing eyes'', outputting the edited image keeping high fidelity to the query image meanwhile well-aligned with the before-after image pair. All other examples in ``addition'', ``removal'', ``style transfer'', and ``replacement'' demonstrate that our \textit{LoC} model consistently generates high-quality edited images that effectively align with before-after image pairs.

\noindent{\bf Comparisons with Competitive Methods.}
We compare the \textit{LoC} model with leading works on image editing with visual instructions. To illustrate the advantages of our method over previous ones, like Analogist and VISII, we compare it with these methods in 6 types of visual instructions. As shown in Figure~\ref{fig:visual_comparisions}, \textit{our LoC model can achieve the desired editing of visual instructions meanwhile keeping good fidelity to the query image.}

Given a before-after image pair that ``a beautiful girl $\to$ a beautiful girl wearing glasses'' and a query image of another girl, \textit{LoC} model outputs an image that the girl in the query image wearing glasses without other changes.
On the other hand, the Analogist method cannot even output photo-realistic images not mention realizing the desired editing. Given a before-after image pair ``a smiling man $\to$ a smiling bald man'' and a query image ``a little boy opening mouth'' the Analogist method fails to edit the image while the VISII method cannot keep good fidelity to the query image. All these comparison examples in Figure~\ref{fig:visual_comparisions} show the superiority of our $\textit{LoC}$ method.

\subsection{Quantitative Evaluation}
\begin{table}[t]
\centering
\caption{Quantitative comparisons on InsturctPix2Pix dataset~\cite{brooks2023instructpix2pix}.}
\label{tab:compare}
\scalebox{0.85}{
\begin{tabular}{lcccc}
\toprule
Methods            & LPIPS $\downarrow$   & Visual CLIP $\uparrow$  & FID $\downarrow$    & Inf. time (s)       \\ 
\midrule
VISII                 & 0.490   & 0.250  &54.09   &423 \\
\midrule
Analogist             & 0.585   & 0.150  &58.14   &\textbf{1}   \\
PromptGIP             & 0.369   & 0.112  &62.51   &\textbf{1}   \\
\textit{LoC} (Ours)   & \textbf{0.289}   &\textbf{0.214}  &\textbf{46.31}   &3   \\
\bottomrule
\end{tabular}}
\label{tab:quantitative_1}
\vspace{-0.1in}
\end{table}
\noindent{\bf LPIPS, Visual CLIP, FID, and Inference Time.}
We demonstrate the advantages of our \textit{LoC} method with quantitative evaluations.
The following metrics are used:
\begin{itemize}
    \item LPIPS to measure the alignment between the edited image and the query image.
    \item Visual CLIP score~\cite{nguyen2023visual} to evaluate how faithfully the editing adheres to the before-after image pair with $cos[(M(B^{'})-M(B)), (M(A^{'})-M(A))]$ where $M$ is the CLIP model. 
    \item Fréchet inception distance (FID) to evaluate the quality of edited images.
    \item Inference time to evaluate the efficiency of models. 
\end{itemize}

For fair comparisons to Analogist, we use open-sourced LLaVA~\cite{liu2023llava} to generate the required text prompt instead of GPT-4V. Following previous work~\cite{gu2024analogist}, we also use a subset of 1000 $<A,A^{'},B>$ data points in InstructPix2Pix dataset to evaluate all the models. 
The experimental results are summarized in Table~\ref{tab:quantitative_1}. Compared with Analogist and PromptGIP methods, we consistently achieve better LPIPS, Visual CLIP, and FID scores using comparable inference time.
VISII method achieves a much better Visual CLIP score than others because it conducts test-time fine-tuning to invert visual instructions into text embeddings and \textit{does not hurt the knowledge of the pre-trained InstructPix2Pix model that is also trained on the InstructPix2Pix dataset}. 

\begin{figure}[t!]
    \centering
\includegraphics[width=0.48\textwidth]{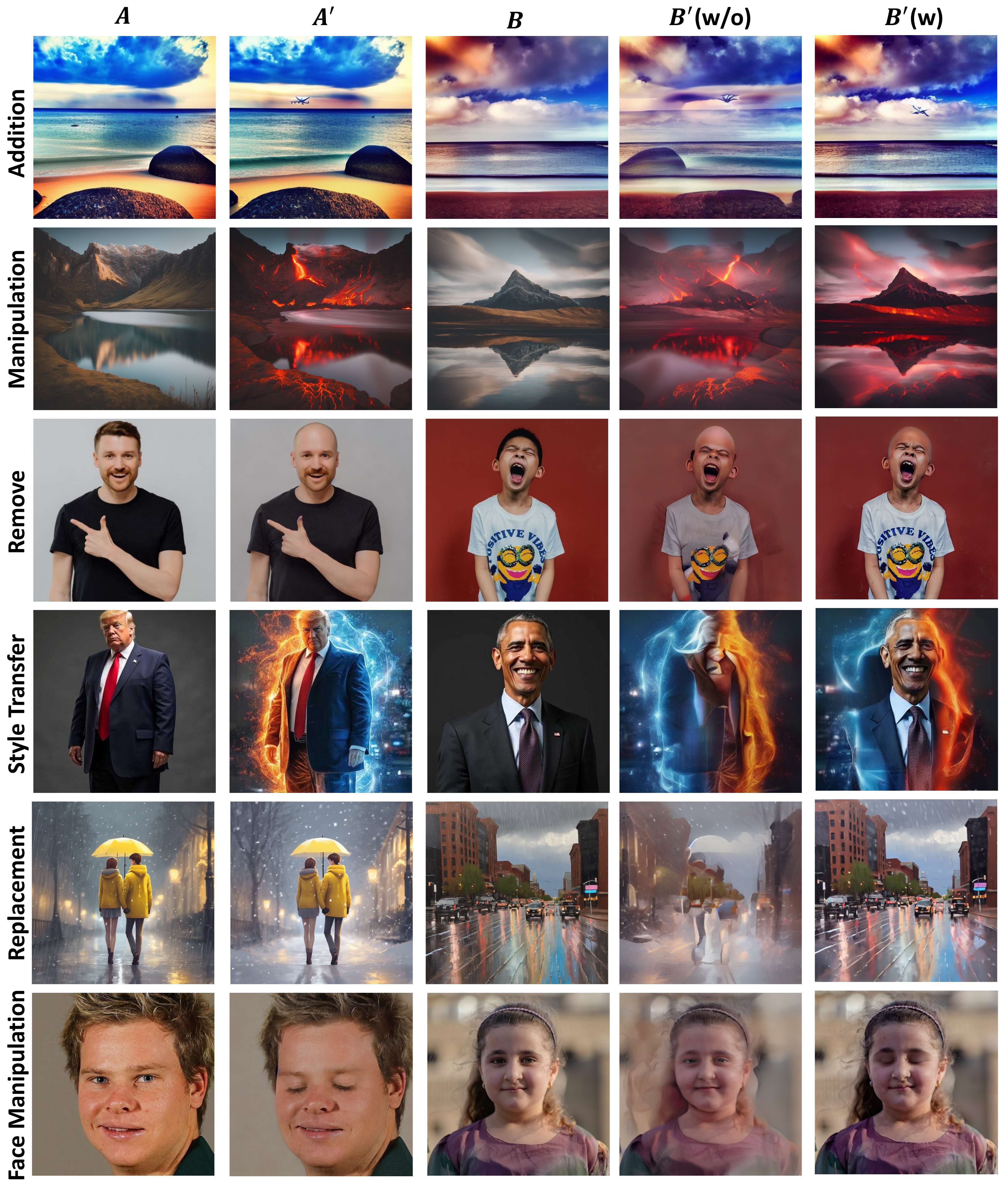} 
    \vspace{-6mm}
    \caption{Ablation on \textit{LoRA Reverse} training.}
    \vspace{-4mm}
\label{fig:ablation_reverse}
\end{figure}

\noindent{\bf User Study.} We conducted a human perceptual evaluation study to validate the effectiveness of our proposed \textit{LoC}. The study consisted of 30 evaluation samples, each including a visual instruction pair, \ie, ($A$, $A'$), and a query image $B$. These samples cover a wide spectrum of editing types, including addition, manipulation, removal, style transfer, replacement, and face manipulation. 
A total of 43 workers from Amazon Mechanical Turk (AMT) participated in the study. For each sample, participants were shown the set ($A$, $A'$, $B$) alongside three edited images 
\begin{wrapfigure}{r}{0.19\textwidth}
    \centering
\includegraphics[width=.19\textwidth]{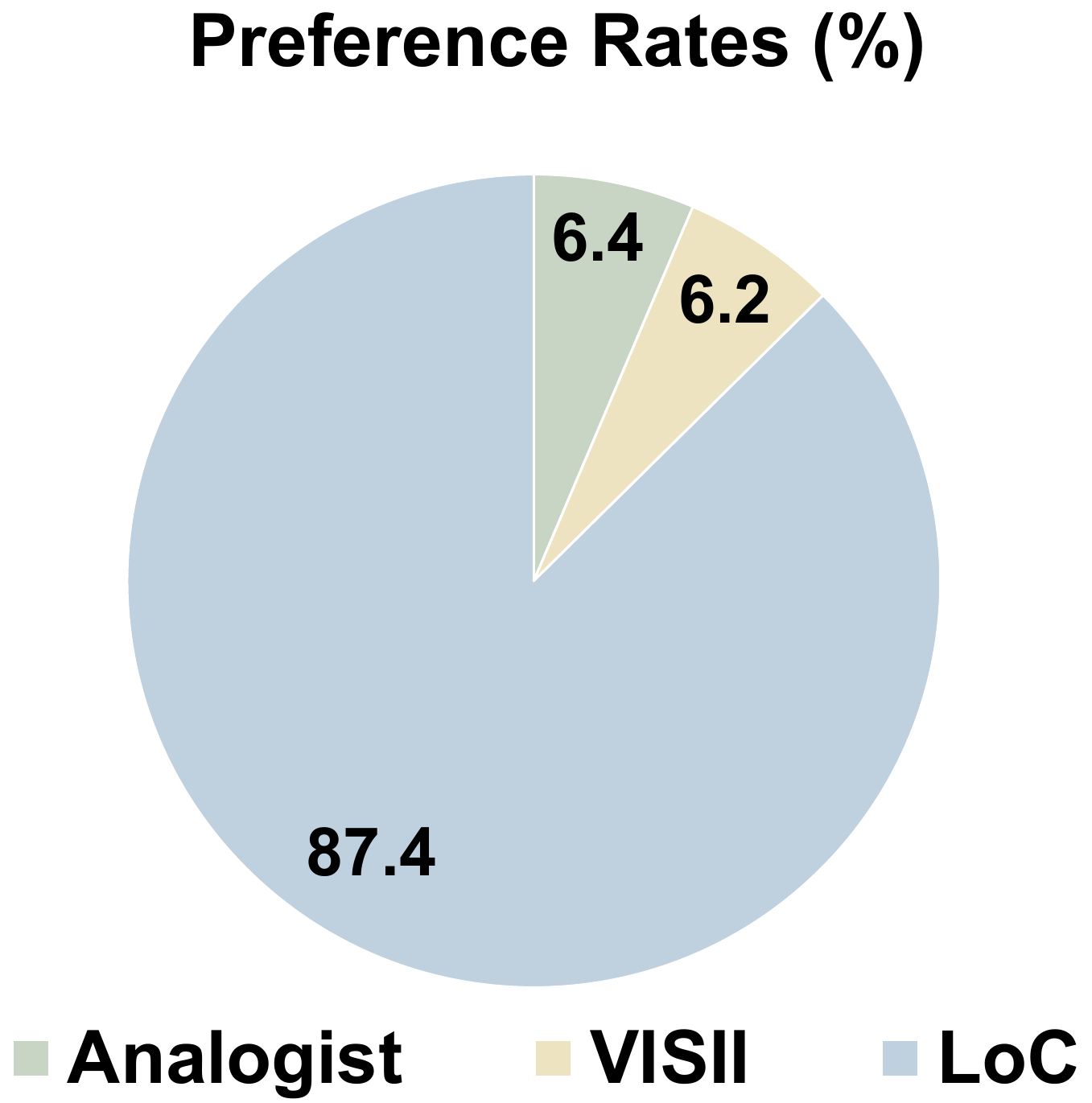}
    \caption{User study statistics.}
    \label{fig:user_study}
    \vspace{-3mm}
\end{wrapfigure}
generated by Analogist, VISII, and our \textit{LoC}, which were randomly shuffled. 
Evaluators were asked to choose the best result based on its similarity to $B$ and its alignment with the visual instruction in ($A$, $A'$). The user study results, shown in Figure \ref{fig:user_study}, highlight the superiority of \textit{LoC} over the other methods, with \textit{LoC} receiving 87.4\% of the preferences.

\begin{figure}[t!]
    \centering
\includegraphics[width=0.5\textwidth]{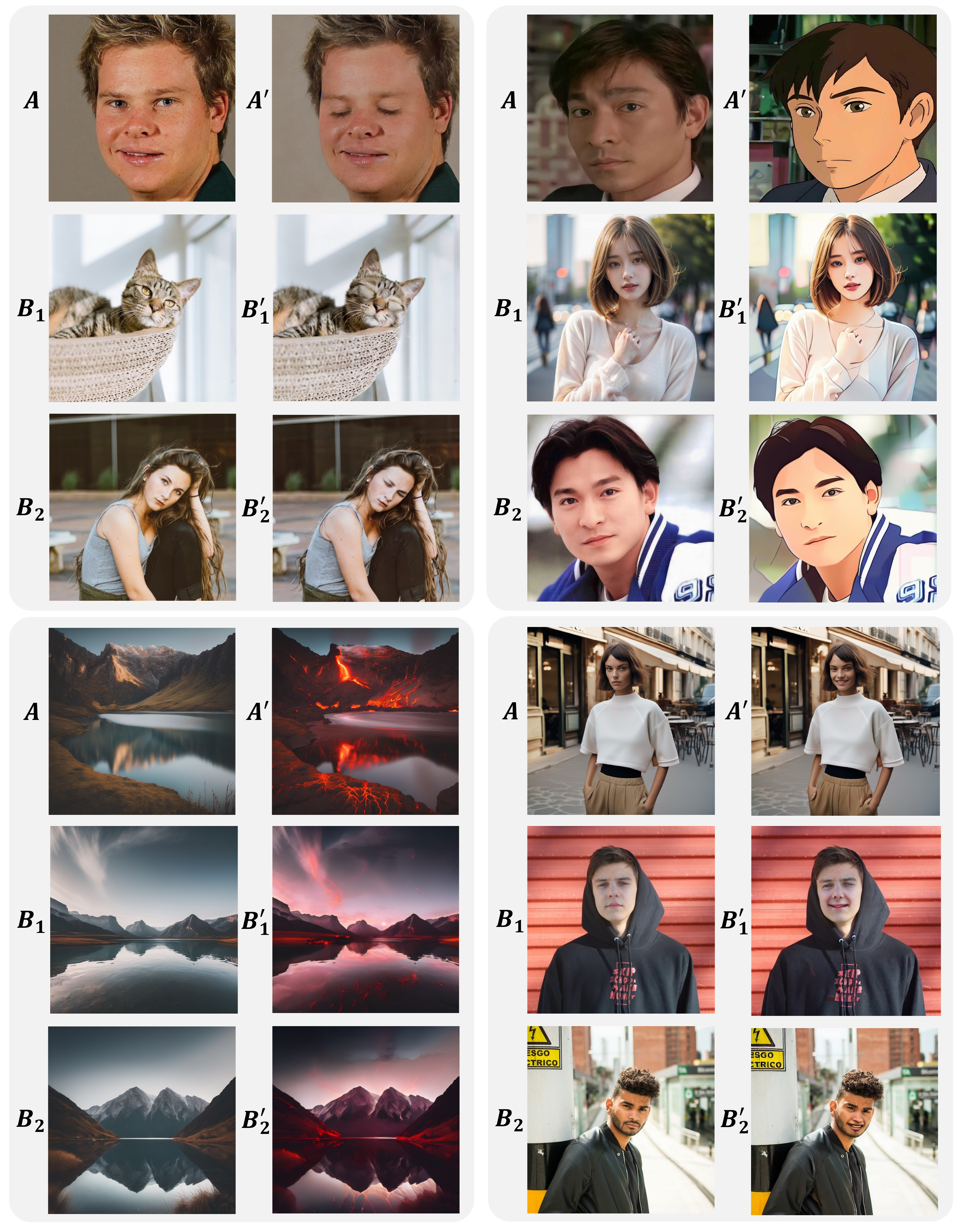} 
    \vspace{-6mm}
    \caption{Ablation on interpretability and reusability. }
    \vspace{-4mm}
\label{fig:ablation_interpretability}
\end{figure}

\begin{figure*}[t!]
    \centering
\includegraphics[width=\textwidth]{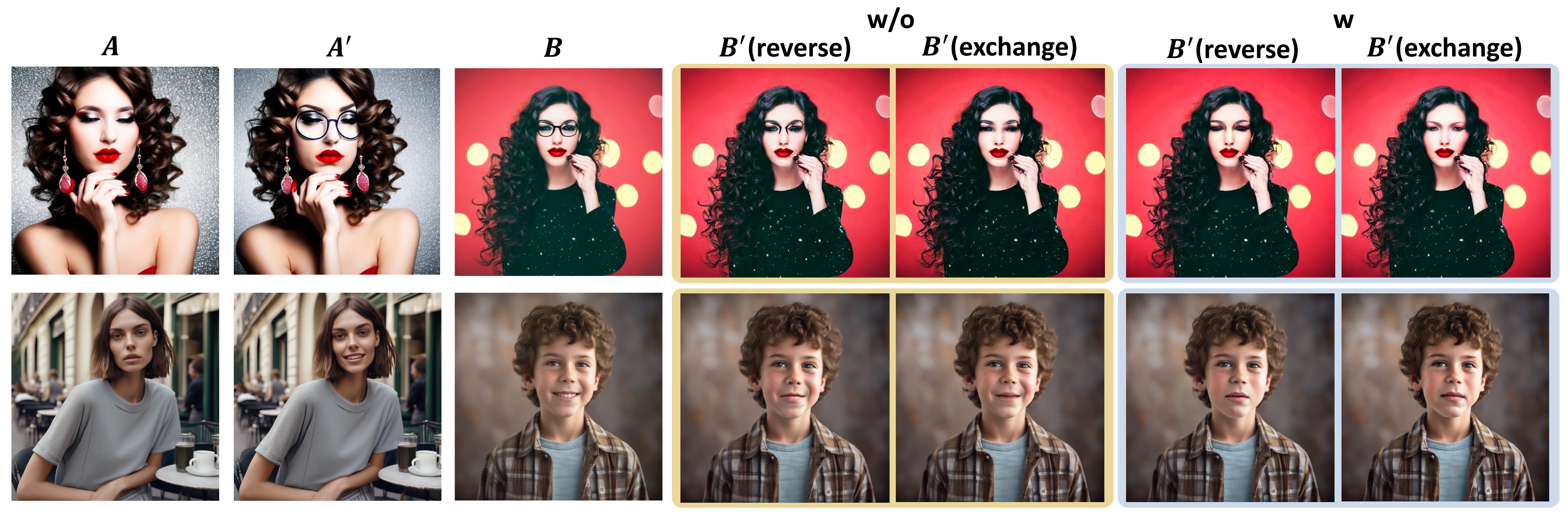} 
    \vspace{-6mm}
    \caption{Ablation on random exchange from $<A,A^{'}>$ to $<A^{'}, A>$ for \textit{LoRA Reverse} training. }
    \vspace{-4mm}
\label{fig:ablation_exchange}
\end{figure*}

\subsection{Ablation Analysis}
\noindent{\bf Ablation on LoRA Reverse Training.}
We conduct ablation on the effectiveness of \textit{LoRA Reverse} training.
Specifically, we train two models w/ and w/o the \textit{LoRA Reverse} training while keeping other experimental settings the same for fair comparisons.
The experimental results are summarized in Figure~\ref{fig:ablation_reverse}.
As illustrated in Figure~\ref{fig:ablation_reverse}, without \textit{LoRA Reverse} training, the appearance leakage issue becomes more pronounced. The edited images $B^{'}$(w/o) closely resemble $A^{'}$ in most cases (\eg, rows 1, 3, and 4 in the figure), suggesting that the learned $\textit{instruction-specific}$ LoRA captures appearance features from the before-after image pairs rather than focusing on the ``change'' between $A$ and $A^{'}$.
In contrast, with \textit{LoRA Reverse} training, the edited images are well-aligned with the before-after pairs and maintain high fidelity to the query images, indicating that the learned \textit{instruction-specific} LoRA can successfully extract the visual instructions while filtering out the appearance information from the before-after image pairs. 
All these examples consistently highlight the effectiveness and necessity of our \textit{LoRA Reverse} technique to large-scale training with \textit{paired data}.

\noindent{\bf Interpretability and Reusability.}
We demonstrate the interpretability and reusability of the learned \textit{instruction-specific} LoRA. With a before-after image pair $<A, A^{'}>$, the generated LoRA $\Delta$ only encodes desired visual instruction, \ie, the ``change'' of $A$ and $A^{'}$. Thus $\Delta$ should be applicable to various query images. As shown in Figure~\ref{fig:ablation_interpretability}, we test the interpretability and reusability of the \textit{instruction-specific} LoRA by editing multiple query images with the same before-after image pair.

Given the before-after image pair ``a smiling man with opened eyes $\to$ the smiling man with closed eyes", the learned LoRA $\Delta$ encodes the instruction of ``make the eyes closed". Applying $\Delta$ to query images of a cat and a girl, both the cat and the girl close their eyes and enjoy a peaceful life in the edited images.
All examples in Figure~\ref{fig:ablation_interpretability} confirm the good interpretability and reusability of the learned \textit{instruction-specific} LoRAs.

\noindent{\bf Random Exchange from $<A,A^{'}>$ to $<A^{'}, A>$ for Consistency in LoRA Reverse.}
We show the effects of random exchange from $<A,A^{'}>$ to $<A^{'}, A>$ during training on model performance.
\textit{LoRA Reverse} can effectively prevent appearance leakage from the before-after image pair $<A, A^{'}>$, enabling large-scale training on \textit{paired data}. However, although we expect $\mathcal{G}(\mathcal{H}(A,A^{'}), B)$ and $\mathcal{G}(-\mathcal{H}(A^{'},A),B)$ can both achieve desired editing with the \textit{LoRA Reverse} training, an inconsistency issue can unexpectedly occur,
\begin{equation}
    \mathcal{H}(A,A^{'}) \neq -\mathcal{H}(A^{'},A).
\end{equation}
Such inconsistency of \textit{instruction-specific} LoRA can potentially degrade the generalization ability of models.
Thus, we randomly exchange the before-after image pair from $<A, A^{'}>$ to $<A^{'}, A>$ during training, implicitly encourage the consistency between $\mathcal{H}(A,A^{'})$ and $-\mathcal{H}(A^{'},A)$. As shown in Figure~\ref{fig:ablation_exchange}, reversing the LoRA, \ie, $-\mathcal{H}(A, A^{'})$, and exchange the before-after image pair, \ie, $\mathcal{H}(A^{'}, A)$, can both achieve successful editing. However, with the random exchange operation for consistency regularization, the edited images keep higher fidelity to the query images and are also more aligned with the before-after image pairs.

\noindent{\bf Ablation on Two-stage Training.}
We show the benefits of the stage-2 finetuning. Our models are trained with a two-stage pipeline: pre-train the models on a large hybrid dataset --- SEED-Data-Edit, then finetune the models on a smaller high-quality dataset --- MagicBrush and part of SEED-Data-Edit. 
We empirically observe that the Visual CLIP score is improved from 0.193 to 0.214 when incorporating stage-2 fine-tuning, indicating its necessity.

\section{Conclusion and Limitation}
In this paper, we propose the framework of \textit{LoRA of Change (LoC)} to address image editing based on visual instructions. With the \textit{LoC} method, we learn to generate \textit{instruction-specific} LoRA weights to encode the ``change'' in the before-after image pair. The disentanglement between \textit{instruction-specific} LoRAs and generation models enhances the interpretability and reusability of our approach. Additionally, the proposed \textit{LoRA Reverse} training can effectively mitigate the problem of appearance leakage from before-after image pairs, which enables large-scale training with only \textit{paired data}. Experimental results demonstrate that our model can support various visual instructions, like addition, removal, manipulation, replacement, and style transfer. Compared with previous methods, the edited images by our model keep better fidelity to query images meanwhile are more aligned with the before-after image pairs.

In line with other image generation models, our model carries potential risks of misuse, such as the generation of harmful content. We are committed to restricting the use of our model strictly for research purposes only.

{
    \small

    \bibliographystyle{ieeenat_fullname}
}

\clearpage
\appendix

\noindent The \textbf{Appendix} is organized as follows:
\begin{itemize}[leftmargin=*]
    \item \textbf{Section~\ref{a}:} gives more details on our proposed \textit{LoC}, \ie, LoRA generation and weight initialization.
    \item \textbf{Section~\ref{b}:} shows more qualitative results, \ie, InsturctPix2Pix dataset and wide spectrum of visual instructions.
\end{itemize}

\section{Additional Details}
\label{a}

\begin{figure}[b]
    \centering
    \includegraphics[width=0.85\linewidth]{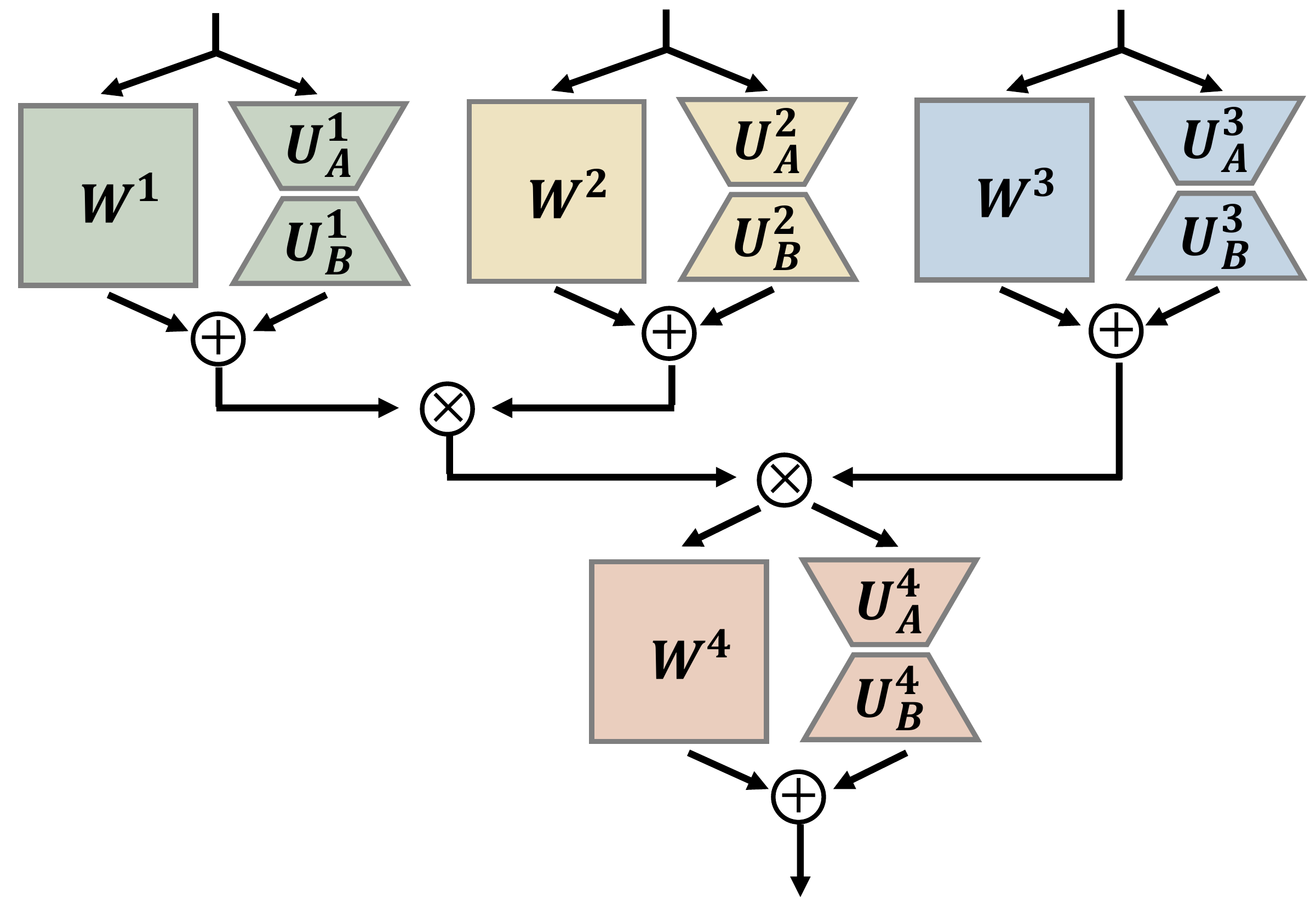}
    \caption{The $i$th self/cross-attention layer of UNet after injecting LoRAs. $W^{1}$, $W^{2}$, $W^{3}$, $W^{4}$ are pre-trained weights of the query, key, value, and output projection layers individually.}
    \label{fig:unet_layer}
\end{figure}

\noindent {\bf LoRA Generation.}
We dynamically generate LoRA weights $\Delta_{i}$ ($i$ is the layer index) for self/cross-attention layers in the UNet of InstructPix2Pix model $\mathcal{G}$.
Following the common practice of LoRA injection in attention layers (\eg, DreamBooth), we apply LoRA on the query, key, value, and output projection layers, demonstrated in Figure~\ref{fig:unet_layer}. 

Specifically, for a linear layer with hidden size $d$, the LoRA weights are $U_A$ and $U_B$, where $U_{A} \in \mathbb{R}^{d \times r}$ and $U_{B} \in \mathbb{R}^{r \times d}$ are low-rank matrices with $r<<d$. Therefore, the generated $\Delta_{i}$ with Eq.~\eqref{eq:i_lora} is split into 8 matrices,
\begin{equation}
    \small
    U_{A}^{1}, U_{A}^{2}, U_{A}^{3}, U_{A}^{4}, U_{B}^{1}, U_{B}^{2}, U_{B}^{3}, U_{B}^{4} \!=\! \mathcal{S}[i](\mathcal{D}(q, f_{vis\_ins})[i]).
\end{equation}
$(U_{A}^{1}, U_{B}^{1})$, $(U_{A}^{2}, U_{B}^{2})$, $(U_{A}^{3}, U_{B}^{3})$, and $(U_{A}^{4}, U_{B}^{4})$
correspond to the query, key, value, and output projection layers.

\noindent {\bf Weight Initialization.}
We empirically observe that a good weight initialization is necessary for fast training convergence. Similar to the original paper~\cite{hu2021lora}, we guarantee $U_{B}^{1}$, $U_{B}^{2}$, $U_{B}^{3}$, and $U_{B}^{4}$ are zeros at the beginning of training.

Specifically, suppose $w_{i} \in \mathbb{R}^{512 \times d}$ is the learnable parameters of the $i$th linear layer in $\mathcal{S}$ ($d$ is the output dimension), the $w_{i}$ is initialized as:
\begin{equation}
    w_{i} = concat(w_{i}^{1}, w_{i}^{2}),
\end{equation}
where $w_{i}^{1} \in \mathbb{R}^{512 \times \frac{d}{2}}$ is randomly initialized, $w_{i}^{2} \in \mathbb{R}^{512 \times \frac{d}{2}}$ is initialized to zero.



\begin{figure}[t]
    \centering
    \includegraphics[height=0.65\textwidth]{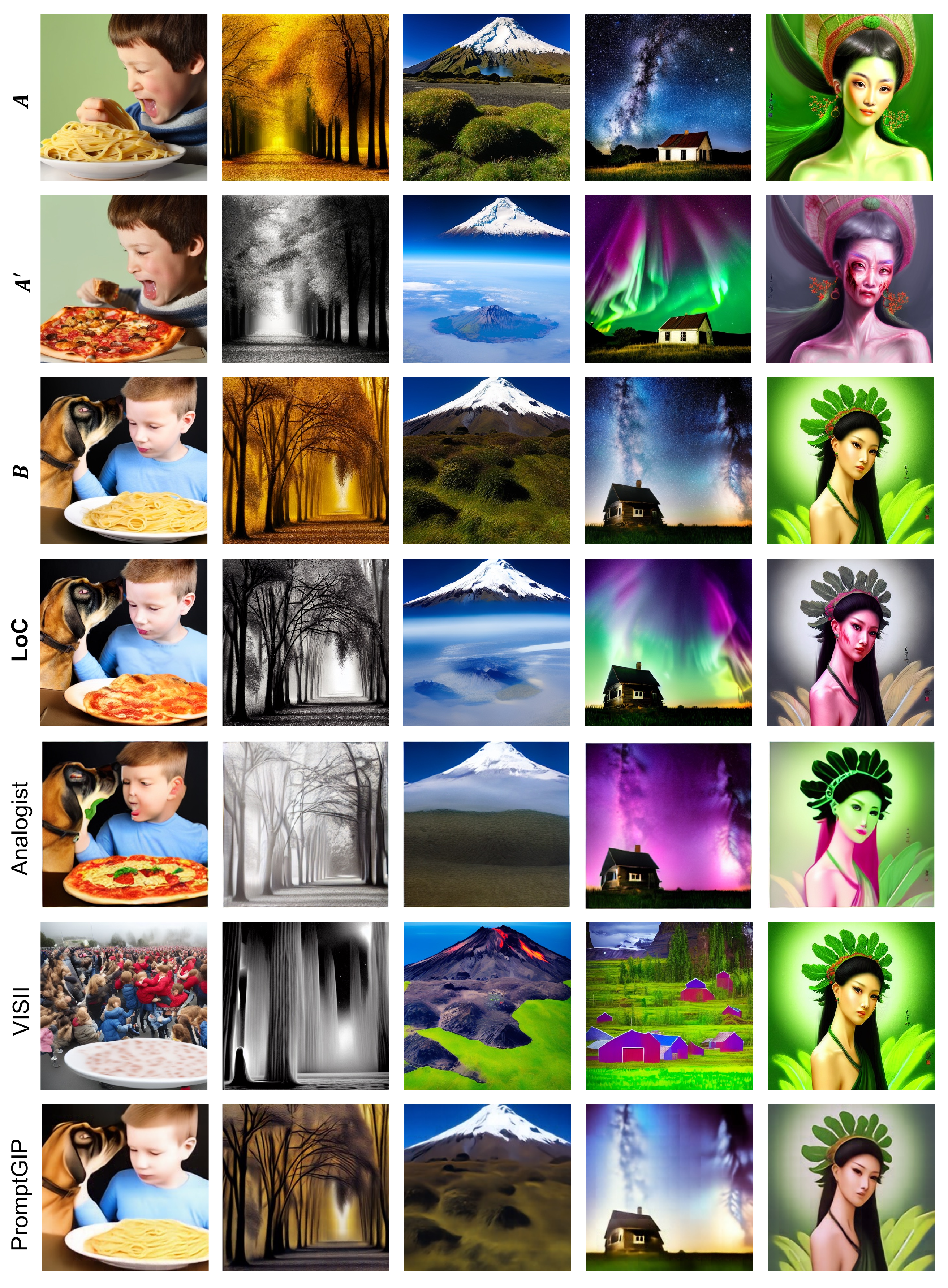}   
    \caption{Qualitative comparison on InstructPix2Pix dataset.}
    \label{fig:ip2p_comparisions_supp}
    \vspace{-0.2cm}
\end{figure}

\begin{figure*}[!htp]
    \centering
    \includegraphics[height=1.0\textheight]{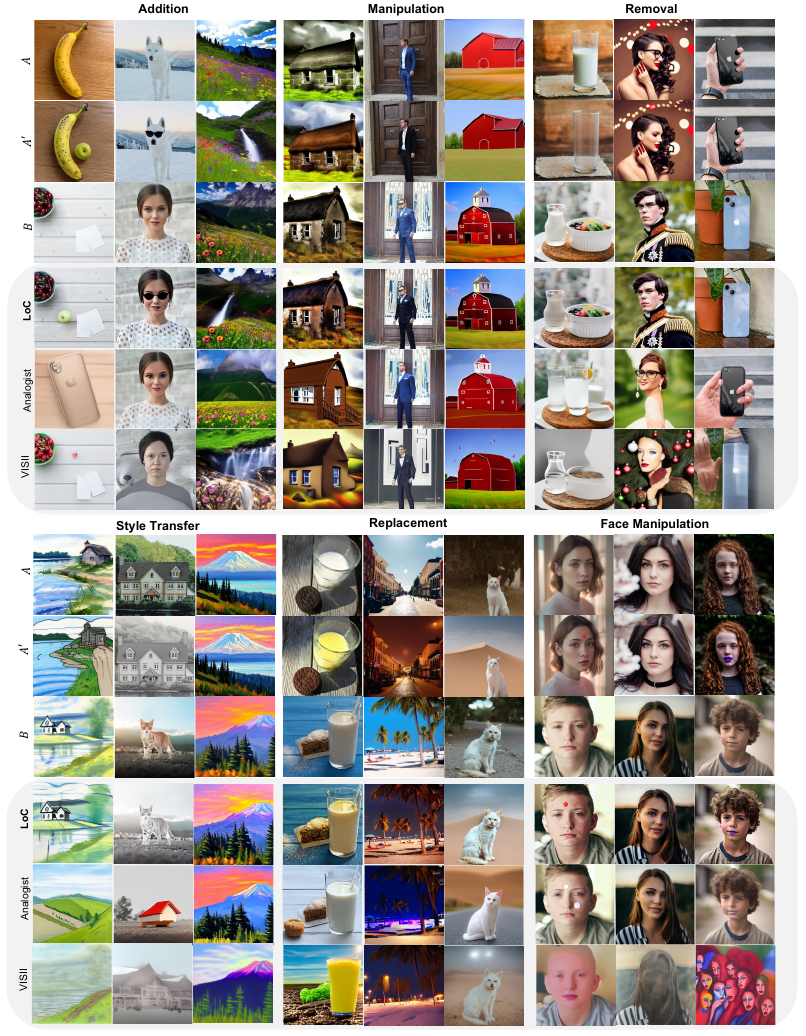}     
    \caption{Comparison of qualitative examples across the 6 editing types between our LoC and two SOTAs.}
\label{fig:visual_comparisions_supp}
\end{figure*}

\section{Additional Qualitative Results}
\label{b}

\noindent \textbf{InstructPix2Pix Dataset.}
The qualitative comparison on the InstructPix2Pix dataset is shown in Figure \ref{fig:ip2p_comparisions_supp}. The edited images with our model enjoy much better quality, keeping higher fidelity to the query images while being more aligned with the before-after image pairs.     

For the example in the first column, the instruction given by $(A, A^{'})$ is replacing pasta with pizza. It could be seen that our \textit{LoC} successfully performs the replace editing while maintaining the fidelity of others in $B$, \eg, the plate. By contrast, although Analogist changes the pasta to pizza, it loses the appearance details of others in $B$, \eg, the face of the boy. VISII neither achieves editability nor maintains fidelity while the edited image from PromptGIP is almost the same as $B$. For other examples, we could change the photograph to black and white, change the Milky Way to the northern lights, and make the lady look like a zombie.

\noindent \textbf{Wide Spectrum of Visual Instructions.} Extra comparison across 6 editing types is listed in Figure \ref{fig:visual_comparisions_supp}. We achieve successful editing by removing the logo on the iPhone, replacing the background of the cat with desert, and putting purple lipstick on the boy, while the other approaches struggle to fulfill editability and fidelity simultaneously.




\end{document}